\documentclass[10pt]{article} 
\usepackage[preprint]{tmlr}

\usepackage{amsmath,amsfonts,bm}

\def\eqref#1{equation~\ref{#1}}

\def\1{\bm{1}}

\DeclareMathAlphabet{\mathsfit}{\encodingdefault}{\sfdefault}{m}{sl}
\SetMathAlphabet{\mathsfit}{bold}{\encodingdefault}{\sfdefault}{bx}{n}

\usepackage{url}
\usepackage{multirow}
\usepackage{graphicx}
\usepackage{booktabs}
\usepackage{caption}
\usepackage{xcolor}
\definecolor{cvprblue}{rgb}{0.21,0.49,0.74}
\usepackage[pagebackref,breaklinks,colorlinks,allcolors=cvprblue]{hyperref}
\usepackage[ruled,vlined,linesnumbered]{algorithm2e}

\title{ESSI: Computationally Efficient Semantic Search for Sparse Autoencoder Concepts}

\author{James Baker\\
UMBC\\
Catonsville, Maryland\\
{\tt\small jbaker15@umbc.edu}
}

\def\month{11}  
\def\year{2026} 
\def\openreview{\url{https://openreview.net/forum?id=XXXX}} 

\begin{document}
\maketitle
\begin{abstract}

Sparse autoencoders (SAEs) decompose diffusion model activations into interpretable features, but steering a model with them requires identifying which features map to a given semantic concept, if any. Existing methods for this concept to feature task are expensive, relying on grid searches or repeated passes through the frozen UNet. We propose Efficient Semantic Sparse Identification (ESSI). Using segmentation masks to label SDXL activations as concept or non-concept, ESSI ranks SAE features and binary-searches for the smallest set on which a linear classifier still separates positive from negative samples while not falling below a performance threshold, then reuses that classifier to calibrate the steering strength. We evaluate the same selected features on both concept insertion and object unlearning. Compared with prior work, ESSI is as cheap or cheaper, and performs well on both insertion and removal.

\end{abstract}    
\section{Introduction}
\label{sec:intro}

Diffusion models \citep{sohldickstein2015deepunsupervisedlearningusing} have emerged as one of the most powerful and commonly used text-to-image models. Since their introduction, they have been made larger \citep{esser2024scalingrectifiedflowtransformers, labs2025flux1kontext}, faster \citep{luo2023latentconsistencymodelssynthesizing}, and higher scoring on metrics of image quality \citep{juggernautxlv9}. In parallel, there has been more interest in interpreting and understanding how these models work. Using a UNet \citep{ronneberger2015unetconvolutionalnetworksbiomedical} as a backbone is very popular, and is a good element to study for interpretability: the middle layers already contain some semantic information \citep{kwon2023diffusionmodelssemanticlatent,baranchuk2022labelefficientsemanticsegmentationdiffusion}, so the activations of these layers are prime targets for using sparse autoencoders \citep{ranzato2006} to disentangle \citep{surkov2024unpackingsdxlturbointerpreting}. We choose to use SDXL, given that its weights, and the weights of a few sparse autoencoders are publicly available.

Sparse representations are not particularly useful if we cannot discover which semantic concepts are encoded within them. A lot of work labels features post hoc \citep{surkov2024unpackingsdxlturbointerpreting,tinaz2025emergence}, taking a feature and then figuring out which human interpretable concepts it matches to. We call this the \textit{feature} \(\rightarrow\) \textit{concept} task. This is different from transforming a human interpretable concept into a good feature, which we call the \textit{concept} \(\rightarrow\) \textit{feature} task. But what constitutes a "good" feature? Interpretability is a less straightforward task than prediction or classification, as the end user's needs are downstream of having an understandable, monosemantic feature for a concept. \citet{gurnee2023findingneuronshaystackcase} looks at language models, and rates features by classification accuracy. We use classification accuracy as a heuristic to find good features, but interpretable features should be judged by how well they can be used to ablate the concept from a model output that would otherwise contain it, or insert the feature into a model output that would otherwise not contain it. To our knowledge, presently there is no work that, starting from a named concept, searches for the features and evaluates the same selected features quantitatively on both removal and insertion in diffusion models.

Additionally, prior work \citep{he2026caslconceptalignedsparselatents,cywinski2025saeuroninterpretableconceptunlearning} for aligning sparse latent dimensions of diffusion models with semantic concepts is also computationally expensive, requiring a grid search over hyperparameters \citep{cywinski2025saeuroninterpretableconceptunlearning}, multiple passes through the frozen UNet \citep{he2026caslconceptalignedsparselatents}, training multiple classifiers, or a complicated convex optimization problem \citep{gurnee2023findingneuronshaystackcase}. Some of these require quadrillions of FLOPs, costing time and energy. Scaling laws \citep{kaplan2020scalinglawsneurallanguage} predict that models will only get larger, so this problem will only get worse. Even with current models, searching for dozens or even hundreds of distinct concepts would be infeasible under the current paradigm. To rectify this, we propose \textit{Efficient Semantic Sparse Identification (ESSI)}. ESSI uses segmentation masks to label each spatial activation as concept or non-concept, ranks SAE features by an importance score, and binary-searches for the smallest feature set on which a linear classifier still reaches 95\% accuracy. The same classifier calibrates the steering strength, so no images are generated during the search. An additional gap in prior work is inadequate comparisons: \citet{cywinski2025saeuroninterpretableconceptunlearning} only scores its method on removal, \citet{he2026caslconceptalignedsparselatents} only scores insertion, and \citet{gurnee2023findingneuronshaystackcase} does neither. We provide a comprehensive comparison of our method against competing methods on both tasks.

\begin{enumerate}
    \item The first comprehensive comparison of \textit{concept} \(\rightarrow\) \textit{feature} methods that evaluates them on both removal and insertion.
    \item Propose ESSI: a new method for the \textit{concept} \(\rightarrow\) \textit{feature} task using between \(\sim\)2000\(\times\) and \(\sim\)9000\(\times\) fewer FLOPs than some of the alternative methods.
    \item Evaluate our method on both insertion and unlearning and show that our method achieves decent scores on removal and insertion with a massive decrease in FLOPs needed.
\end{enumerate}

\section{Related Work}
\label{sec:related}

\subsection{Mechanistic Interpretability}

Machine learning models are often described as "black boxes". It's not clear what human-understandable concepts are actually represented by hidden activations. Occasionally, we get lucky, such as when early layers of the Inception model \citep{szegedy2014goingdeeperconvolutions} turned out to be Gabor filters \citep{olah2020overview}, but this is the exception, not the norm. One of the major complications is that hidden activations are usually polysemantic \citep{scherlis2025polysemanticitycapacityneuralnetworks}, so one neuron can activate on seemingly unrelated concepts. The sparse autoencoder (SAE) \citep{ranzato2006,ng2011sparse} was proposed to rectify this. SAEs are shallow autoencoders, where the latent layer is of a significantly higher dimensionality, and a sparsity penalty (such as \(L_0\)) is applied to the latents. Some common variants include Jump ReLU \citep{rajamanoharan2024improvingdictionarylearninggated}, Top K \citep{gao2024scalingevaluatingsparseautoencoders}, and Matryoshka \citep{bussmann2025learningmultilevelfeaturesmatryoshka}. SAEs have been applied to increasingly larger NLP architectures, such as a single transformer block \citep{bricken2023monosemanticity}, a small GPT \citep{bloom2024gpt2residualsaes}, and even Claude \citep{templeton2026scalingmonosemanticityextractinginterpretable}. SAE features could be used for steering \citep{arad-etal-2025-saes,chen2025personavectorsmonitoringcontrolling} these models, with implications for AI safety \citep{hua2026steeringevaluationawarelanguagemodels}. We adopt the \textit{sparse coding} approach of \citet{gurnee2023findingneuronshaystackcase}, but deviate from their approach by studying a vision model, not a text model and allowing for \(k \geq 1\).

\subsection{SAEs for Vision}
SAEs have also been used post hoc on trained vision foundation models, such as CLIP \citep{pachsparse2025} and Vision Transformers \citep{stevens2025interpretabletestablevisionfeatures,stevens2025openendedvisualscientificdiscovery}, as well as incorporated into the training of autoencoders \citep{kulkarni2026interpretablesteerableconceptbottleneck}. For text-to-image models, SAEs have been applied to the text embeddings of the conditioning prompts \citep{he2025singleneuronworksprecise,kim2025conceptsteerersleveragingksparse} as well as the activations of the internal layers of models such as SDXL \citep{podell2023sdxlimprovinglatentdiffusion,surkov2025onestepenoughsparseautoencoders}. Some approaches like \citet{tinaz2025emergence} analyze the sparse features and then assign, post hoc, semantic labels to each latent dimension. However, doing the inverse, of starting with a semantic concept and then finding the matching features, is nontrivial. So far there are four approaches:
\begin{itemize}
    \item SAeUron \citep{cywinski2025saeuroninterpretableconceptunlearning} scores features by how much more they activate on concept data than on other data, then grid-searches the number of features and the removal strength by generating images and scoring them with an unlearning metric.
    \item Concept Aligned Sparse Latents (CASL) \citep{he2026caslconceptalignedsparselatents} trains a concept-aligned mapping on top of the SAE, with a loss that needs passes through the UNet and CLIP.
    \item Adaptive Thresholding \citep{gurnee2023findingneuronshaystackcase} trains a series of classifiers that iteratively decrease the number of features, where at each step the probe is retrained to use the top neurons from the last step.
    \item Optimal Sparse Probing (OSP) \citep{Bertsimas_2021_osp} formulates feature selection as a binary optimization problem, and uses a cutting-plane algorithm to estimate the optimal subset of features.
\end{itemize}

\subsection{Unlearning}
One of the reasons the \textit{concept} \(\rightarrow\) \textit{feature} task is more important than the \textit{feature} \(\rightarrow\) \textit{concept} task is that specific concepts need to be removed from models. Modern models are trained on large datasets full of unsavory concepts \citep{birhane2021multimodaldatasetsmisogynypornography}, so making them unlearn \citep{nguyen2024surveymachineunlearning} is useful for both text \citep{geng2025comprehensivesurveymachineunlearning} and image models \citep{xie2025erasingconceptssteeringgenerations}. There is a wealth of methods that are used to ablate controversial concepts from generative image models \citep{xing2024efuf,cheng2024multidelete,fan2024salun,heng2023selective,gandikota2023erasing,wu2024erasediff}. These approaches sometimes utilize extra weights and adapters \citep{huang2024recelerreliableconcepterasing, gao2025eraseanythingenablingconcepterasure,zhang2025steerdiffsteeringsafetexttoimage}, some sort of adversarial training \citep{nie2025erasing,zhang2024defensiveunlearningadversarialtraining} or a new optimization objective \citep{li2025setstraightautosteeringdenoising,kim2026trainingfreesafedenoiserssafe}.

\section{Methodology}
\label{sec:methodology}

Given a concept \(c\) (e.g.\ an object), ESSI finds a small set of sparse autoencoder (SAE) features at a U-Net layer that represent \(c\), a removal strength \(\gamma_c\) that suppresses the concept when those features are rescaled at inference time, and an insertion strength \(\alpha_c\) that adds the concept to an image by adding the same features' decoder directions.
Like SAeUron \citep{cywinski2025saeuroninterpretableconceptunlearning}, ESSI ranks features by how much more activation they carry on images of \(c\) than on images of other concepts. It differs from SAeUron in two ways.
(i) We use segmentation masks throughout. The ranking is computed only over the segmented region of each image, so background patches contribute to neither the concept's nor the other concepts' statistics. The probes that choose the number of features and \(\gamma_c\) contrast the patches inside and outside the mask \emph{within the same images}, which share the same prompt, style and background.
(ii) Instead of keeping a fixed number of top-ranked features with a hand-tuned strength, we choose the number of features \(k\), \(\gamma_c\) and \(\alpha_c\) automatically, using linear probes on cached activations, without generating any images for the search.
The full procedure is given in Algorithm~\ref{alg:essi}; we describe each step below.

\paragraph{Notation.}
Let \(\mathcal{L}\) be the set of candidate layers, \(\mathcal{C}\) the set of concepts (\(c \in \mathcal{C}\)), \(T_{c'}\) the prompts of concept \(c' \in \mathcal{C}\), and \(E_l:\mathbb{R}^{D_l}\to\mathbb{R}^{F}\) the SAE encoder at layer \(l\), with \(F \gg D_l\) latent features and decoder \(W^{\text{dec}}_l \in \mathbb{R}^{D_l \times F}\).
We write \(z_i\) for the \(i\)-th coordinate of an SAE code \(z = E_l(a)\).

\subsection{Collecting activations}
\label{sec:method:activations}
For every concept \(c' \in \mathcal{C}\), we build the prompt set \(T_{c'}\) from the prompt templates of \citet{ruiz2023dreamboothfinetuningtexttoimage}, filling the subject slot with \(c'\) (e.g.\ ``a \(c'\) in the jungle'').
We generate one image per prompt with SDXL-Turbo \citep{sauer2023adversarialdiffusiondistillation}, a distilled version of SDXL \citep{podell2023sdxlimprovinglatentdiffusion}, using a single denoising step.
Following \citet{surkov2024unpackingsdxlturbointerpreting}, we record the residual update of the transformer block, i.e.\ the block's output minus its input.
For layer \(l\) and prompt \(t\), this gives an activation map \(a_{l,t} \in \mathbb{R}^{H_l \times W_l \times D_l}\), whose entry \(a_{l,t}[x,y] \in \mathbb{R}^{D_l}\) we call a \emph{patch}.
We encode every patch with the pretrained TopK SAEs of \citet{surkov2024unpackingsdxlturbointerpreting}, \(z_{l,t}[x,y] = E_l(a_{l,t}[x,y])\).

\subsection{Segmentation masks}
\label{sec:method:masks}
For each image generated from a prompt \(t \in T_{c'}\), we prompt SAM~3 \citep{carion2026sam3segmentconcepts} with the text \(c'\) and take the union of the returned instance masks; if SAM3 returns no mask, the mask is empty.
We downsample the pixel mask to each layer's resolution by majority vote: a patch is inside the mask if more than half of its pixels are. This gives \(m_{l,t} \in \{0,1\}^{H_l \times W_l}\).
For the concept \(c\), the patches of its own images are split into
\begin{equation}
\begin{split}
    \mathcal{A}^{+}_{l,c} = \bigcup_{t \in T_c} \{\, a_{l,t}[x,y] : m_{l,t}[x,y] = 1 \,\}, \\
     \mathcal{A}^{-}_{l,c} = \bigcup_{t \in T_c} \{\, a_{l,t}[x,y] : m_{l,t}[x,y] = 0 \,\}
    \end{split}
    \end{equation}
with SAE codes \(\mathcal{D}^{\pm}_{l,c} = \{ E_l(a) : a \in \mathcal{A}^{\pm}_{l,c} \}\). These are the positives and negatives of the probes in Sections~\ref{sec:method:k} and~\ref{sec:method:removal}; an image with an empty mask contributes only negatives.

\subsection{Ranking features}
\label{sec:method:score}
We use the feature-importance score of \citet{cywinski2025saeuroninterpretableconceptunlearning}, computed on masked regions only.
For an image \(t\) with a non-empty mask, let
\begin{equation}
    \bar{z}_{l,t} = \frac{1}{|\{(x,y) : m_{l,t}[x,y] = 1\}|} \sum_{(x,y) :\, m_{l,t}[x,y] = 1} z_{l,t}[x,y]
\end{equation}
be its mean code over the mask (images with an empty mask are left out of the ranking).
For a set of images \(T'\), let \(\mu_i(T') = \frac{1}{|T'|} \sum_{t \in T'} (\bar{z}_{l,t})_i\), and let \(T_{\neg c} = \bigcup_{c' \in \mathcal{C} \setminus \{c\}} T_{c'}\) be the images of the other concepts.
Each feature is scored by how much more of the total activation mass it takes up on the concept's regions than on the other concepts' regions:
\begin{equation}
    \text{score}(i,l,c) =
    \frac{\mu_i(T_c)}{\sum_{j=1}^{F} \mu_j(T_c) + \delta}
    - \frac{\mu_i(T_{\neg c})}{\sum_{j=1}^{F} \mu_j(T_{\neg c}) + \delta},
\end{equation}
where \(\delta\) is a small constant for numerical stability.
SAeUron computes the same score with \(\bar{z}_{l,t}\) averaged over \emph{every} patch of the image; restricting it to the mask removes the background, which is shared across concepts, from both terms.
Let \(S_{l,c} = (s_1,\dots,s_n)\) be the features with a positive score, sorted by descending score, and let \(S_{l,c}[{:}k] = (s_1,\dots,s_k)\) be the top \(k\).

\subsection{Choosing the number of features}
\label{sec:method:k}
We hold out 20\% of the concept's images (split by image, so that patches from one image never appear in both splits) and use the rest for training.
For a feature subset \(I\), let \(P_{l,c}(I)\) be an \(\ell_2\)-regularized logistic regression trained to separate \(\mathcal{D}^{+}_{l,c}\) from \(\mathcal{D}^{-}_{l,c}\) using only the coordinates \(z_I\), with each feature rescaled to a maximum of 1.
Let \(\text{acc}(\cdot)\) be its accuracy on the held-out patches.
We select the smallest \(k\) whose probe reaches 95\% of the accuracy of the probe trained on every feature:
\begin{equation}
\begin{split}
    k_{l,c} = \min \bigl\{\, k \le K : {} &\text{acc}\bigl(P_{l,c}(S_{l,c}[{:}k])\bigr) \\
    &\ge 0.95 \cdot \text{acc}\bigl(P_{l,c}(\{1,\dots,F\})\bigr) \,\bigr\}
\end{split}
\end{equation}
with \(K = \min(64, |S_{l,c}|)\); if no \(k \le K\) reaches the target, we use \(k_{l,c} = K\).
We find \(k_{l,c}\) by binary search over \(k \in \{1,\dots,K\}\). This assumes accuracy does not decrease as \(k\) grows; we log every \(k\) the search evaluates so that the assumption can be checked.
The selected features are \(I_{l,c} = S_{l,c}[{:}k_{l,c}]\).
We do not propose a method for choosing the layer: ESSI is run at each layer \(l \in \mathcal{L}\) independently.

\subsection{Removal}
\label{sec:method:removal}
Following \citet{cywinski2025saeuroninterpretableconceptunlearning}, we remove \(c\) by rescaling the selected features in proportion to how strongly they fire on the concept.
Let \(\bar{\mu}_i\) be the mean activation of feature \(i\) over every patch of the concept's images, i.e.\ over \(\mathcal{D}^{+}_{l,c} \cup \mathcal{D}^{-}_{l,c}\) (patches where the feature is inactive count as zero).
The edited code is
\begin{equation}
    \tilde{z}_i =
    \begin{cases}
        \gamma_c\, \bar{\mu}_i\, z_i & \text{if } i \in I_{l,c}, \\
        z_i & \text{otherwise},
    \end{cases}
\end{equation}
so \(\gamma_c = 0\) zeroes the features and \(\gamma_c < 0\) pushes them in the opposite direction.
Unlike SAeUron, we do not restrict the edit to patches where a feature exceeds a threshold: it is applied at every patch, and leaves patches where a selected feature is inactive unchanged.
To keep the SAE's reconstruction error out of the edit, we do not replace the block's output with the decoded edited code. Instead, we subtract only the change that the selected features contribute:
\begin{equation}
    h' = h - W^{\text{dec}}_{l}\,(z - \tilde{z}),
\end{equation}
where \(h\) is the output of block \(l\) and \(z = E_{l}(h - h_{\text{in}})\) is the code of its residual update.

\paragraph{Choosing \(\gamma_c\).}
To choose \(\gamma_c\) without generating any images, we train a dense probe \(P^{\text{dense}}_{l,c}\) (feature standardization followed by \(\ell_2\)-regularized logistic regression) on the raw activations \(\mathcal{A}^{\pm}_{l,c}\) of the training images.
Because this probe reads the activations directly rather than the SAE codes, it can detect the concept even through features outside \(I_{l,c}\).
On the held-out images, let \(\mathcal{H}\) be the set of positive patches that \(P^{\text{dense}}_{l,c}\) classifies as positive before the edit.
We apply the edit offline, exactly as it is applied during generation, and choose the weakest \(\gamma_c \in [-50, 0]\) after which at most 5\% of \(\mathcal{H}\) is still classified as positive:
\begin{equation}
\begin{split}
    \gamma_c = \max \Bigl\{\, \gamma \in [-50, 0] : {} &\tfrac{1}{|\mathcal{H}|}\,\bigl|\{\, a \in \mathcal{H} : P^{\text{dense}}_{l,c}(a'_\gamma) = 1 \,\}\bigr| \\
    &\le 0.05 \,\Bigr\},
\end{split}
\end{equation}
where \(a'_\gamma\) is patch \(a\) after the edit with strength \(\gamma\).
We find \(\gamma_c\) by bisection (12 steps). If even \(\gamma = -50\) leaves more than 5\% of \(\mathcal{H}\) positive, we use \(\gamma_c = -50\).
This check only covers block \(l\); later blocks and the text conditioning can restore the concept, so we measure removal on generated images in Section~\ref{sec:experiments}.

\subsection{Insertion}
\label{sec:method:insertion}
The same features can also add \(c\) to an image that does not contain it.
We take base images generated from a fixed prompt with a different subject \(s \neq c\) (e.g.\ another object), and regenerate each with the same prompt and seed while adding the selected features' decoder directions inside the base subject's region.
Let \(M_s\) be the SAM~3 mask of \(s\) in the unedited base image, downsampled to layer \(l\) by majority vote, and let \(\nu_i\) be the mean activation of feature \(i\) over the SAE's training data.
At every patch \((x,y)\) with \(M_s[x,y] = 1\), the block's output becomes
\begin{equation}
    h'[x,y] = h[x,y] + \alpha_c\, W^{\text{dec}}_{l}\, v, \qquad
    v_i = \begin{cases} \nu_i & \text{if } i \in I_{l,c}, \\ 0 & \text{otherwise,} \end{cases}
\end{equation}
and patches outside the mask are left unchanged.
Unlike removal, the edit does not depend on the block's own code: only the direction of the selected features is added.

\paragraph{Choosing \(\alpha_c\).}
We consider two choices.
The first reuses the removal strength, \(\alpha_c = -\gamma_c\), at no extra cost.
Since \(\gamma_c\) multiplies \(\bar{\mu}_i z_i\) while \(\alpha_c\) multiplies \(W^{\text{dec}}_l v\), this reuses the number rather than the size of the edit.
The second chooses \(\alpha_c\) the way \(\gamma_c\) is chosen, without generating any images.
We train a dense probe \(Q_{l,c}\) (feature standardization followed by \(\ell_2\)-regularized logistic regression) on the raw activations \(\mathcal{A}^{\pm}_{l,c}\) of \emph{all} of the concept's images; no images need to be held out, because the probe is evaluated on the base images.
Let \(\mathcal{B}_c\) be the patches of the cached, unedited base-image activations at layer \(l\) that lie inside the base subjects' masks, over the base images whose subject is not \(c\).
We add the injection offline, exactly as it is added during generation, and choose the weakest strength after which at least 95\% of those patches look like \(c\) to the probe:
\begin{equation}
\begin{split}
    \alpha_c = \min \Bigl\{\, \alpha \in [0, 50] : {} &\tfrac{1}{|\mathcal{B}_c|}\,\bigl|\{\, b \in \mathcal{B}_c : {} \\
    &Q_{l,c}(b + \alpha\, W^{\text{dec}}_l v) = 1 \,\}\bigr| \ge 0.95 \,\Bigr\}.
\end{split}
\end{equation}
We find \(\alpha_c\) by bisection (12 steps), which assumes that the share of positive patches does not decrease as \(\alpha\) grows.
If even \(\alpha = 50\) does not reach 95\%, we use \(\alpha_c = 50\); if the patches already reach it without any injection, we inject nothing.
As with \(\gamma_c\), this check only covers block \(l\), so we measure insertion on the generated images in Section~\ref{sec:experiments}.
\begin{algorithm}[t]
\caption{ESSI feature selection, removal strength and insertion strength for concept \(c\) at layer \(l\)}
\label{alg:essi}
\KwIn{concept \(c\), concepts \(\mathcal{C}\), prompt templates, layer \(l\), SAE \(E_l\), base images with subject masks \(M_s\)}
\KwOut{\(I_{l,c}, \gamma_c, \alpha_c\)}
For every \(c' \in \mathcal{C}\): generate one image per prompt in \(T_{c'}\) with SDXL-Turbo (1 step); record residual updates \(a_{l,t}\) and encode them\;
Segment \(c'\) in each image of \(T_{c'}\) with SAM~3; downsample to \(m_{l,t}\) by majority vote\;
Rank features by the masked score\((\cdot,l,c)\) (concept's regions vs.\ the other concepts' regions) \(\to S_{l,c}\)\;
Split the patches of \(T_c\) into \(\mathcal{A}^{\pm}_{l,c}\) by the mask; encode to \(\mathcal{D}^{\pm}_{l,c}\)\;
Split the images of \(T_c\) 80/20 into train / held-out\;
Binary-search the smallest \(k_{l,c} \le \min(64, |S_{l,c}|)\) with \(\text{acc}(P_{l,c}(S_{l,c}[{:}k])) \ge 0.95\,\text{acc}(P_{l,c}(\text{all}))\); \(I_{l,c} \gets S_{l,c}[{:}k_{l,c}]\)\;
Train dense probe \(P^{\text{dense}}_{l,c}\) on \(\mathcal{A}^{\pm}_{l,c}\) (train split)\;
Bisect \(\gamma_c \in [-50,0]\): weakest \(\gamma\) leaving \(\le 5\%\) of held-out detected positives positive (\(-50\) if none does)\;
Train dense probe \(Q_{l,c}\) on \(\mathcal{A}^{\pm}_{l,c}\) (all images); collect the masked base patches \(\mathcal{B}_c\) (bases with \(s \neq c\))\;
Bisect \(\alpha_c \in [0,50]\): weakest \(\alpha\) after which \(\ge 95\%\) of \(\mathcal{B}_c + \alpha W^{\text{dec}}_l v\) is classified positive (\(50\) if none does); or set \(\alpha_c = -\gamma_c\)\;
\KwRet{\(I_{l,c}, \gamma_c, \alpha_c\)}\;
\end{algorithm}
\section{Experiments}
\label{sec:experiments}
We learn features for the 20 object concepts of the UnlearnCanvas benchmark \citep{zhang2024unlearncanvas}.
Images are classified with the UnlearnCanvas object and style classifiers \(f_{\text{obj}}\) and \(f_{\text{sty}}\) (ViT-L/16). We retrain both with the original architecture and training recipe, but on images generated by SDXL-Turbo, the diffusion model we edit, since the released classifiers had trouble distinguishing styles on its images (Appendix~\ref{appendix:cra}).
Unless stated otherwise, all features are learned on the last downsampling block before the bottleneck, \texttt{down.2.1}; preliminary experiments showed that features learned at this block outperform those of the other blocks (Appendix~\ref{appendix:block}).
The exception is CASL, whose authors use the bottleneck block, so we report CASL at both \texttt{down.2.1} (CASL (down)) and the bottleneck \texttt{mid.0} (CASL (mid)).
CASL is trained for 5k steps.
Implementation and training details for all methods are in Appendix~\ref{appendix:implementation}.

\begin{figure}
    \centering
    \includegraphics[scale=0.25]{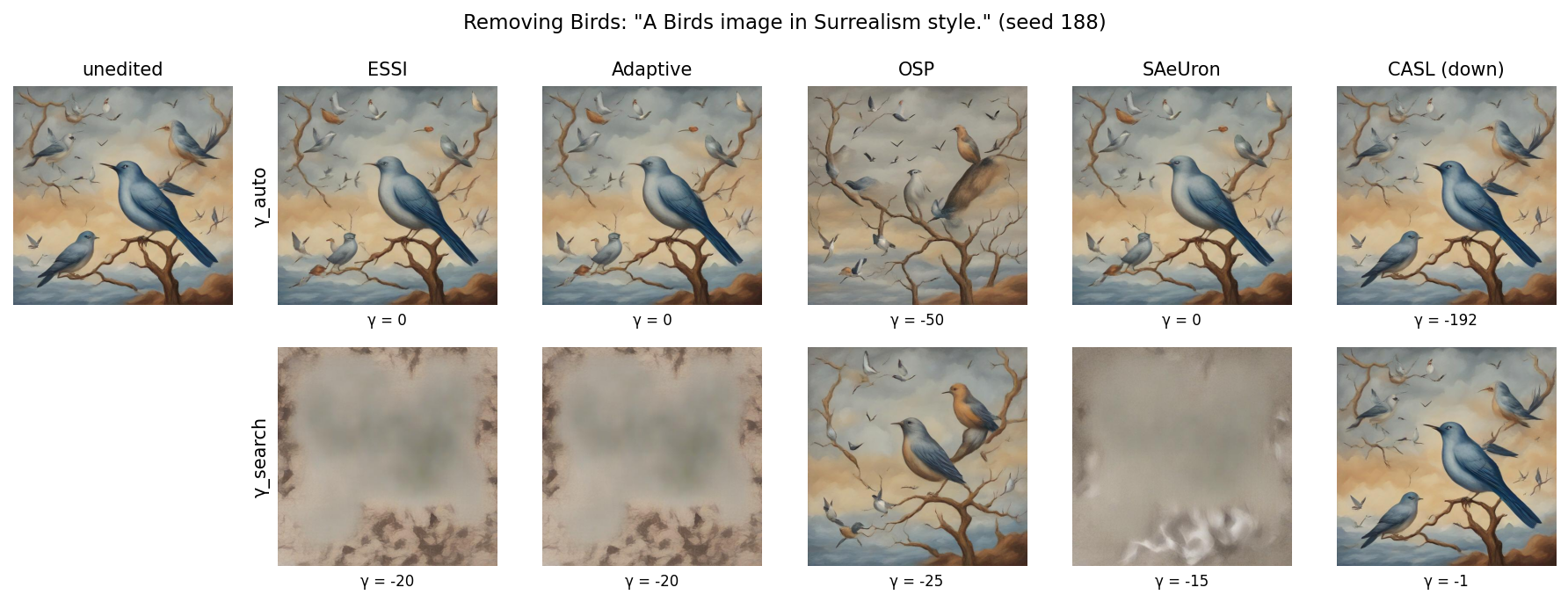}
    \includegraphics[scale=0.25]{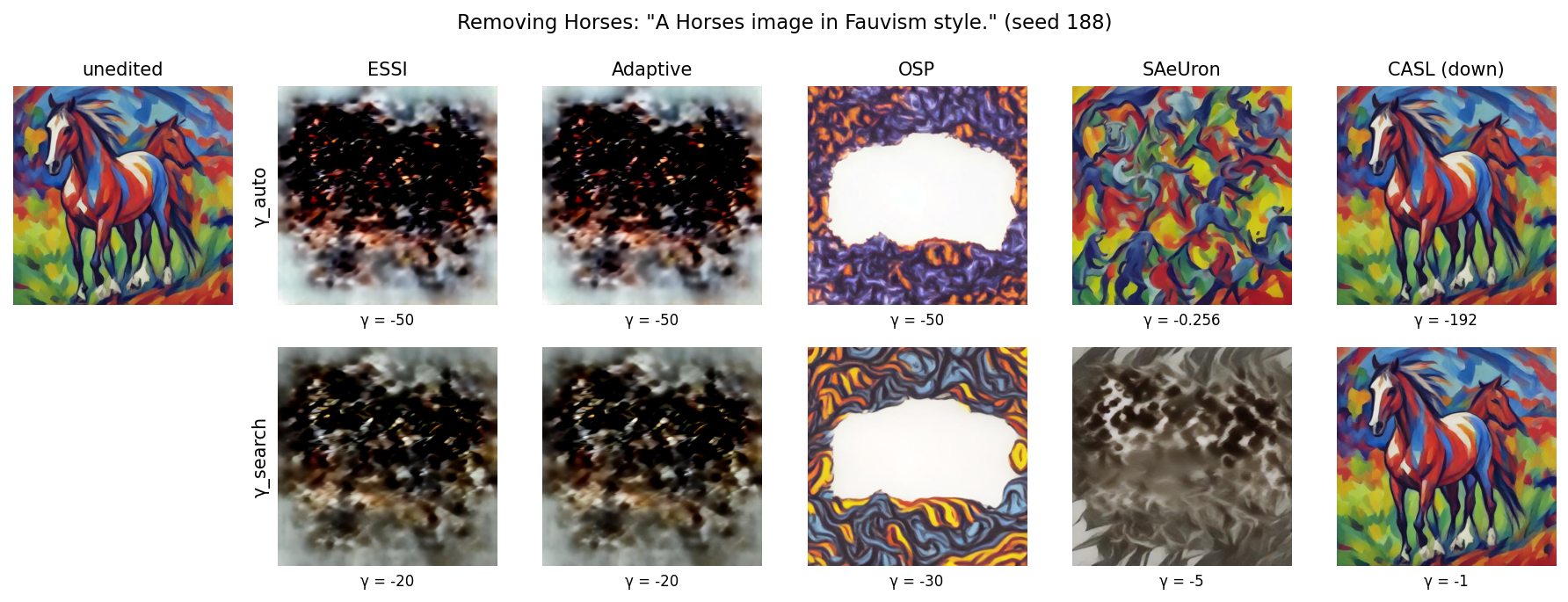}
    \caption{Concept removal examples at \texttt{down.2.1}. Each panel shows the unedited image (left) and each method's edit with \(\gamma_{\text{auto}}\) (top row) and \(\gamma_{\text{search}}\) (bottom row). Top: removing \emph{Birds} from ``A Birds image in Surrealism style.'' Bottom: removing \emph{Horses} from ``A Horses image in Fauvism style.''}\label{fig:removal}
\end{figure}

\begin{figure}
    \centering
    \includegraphics[scale=0.25]{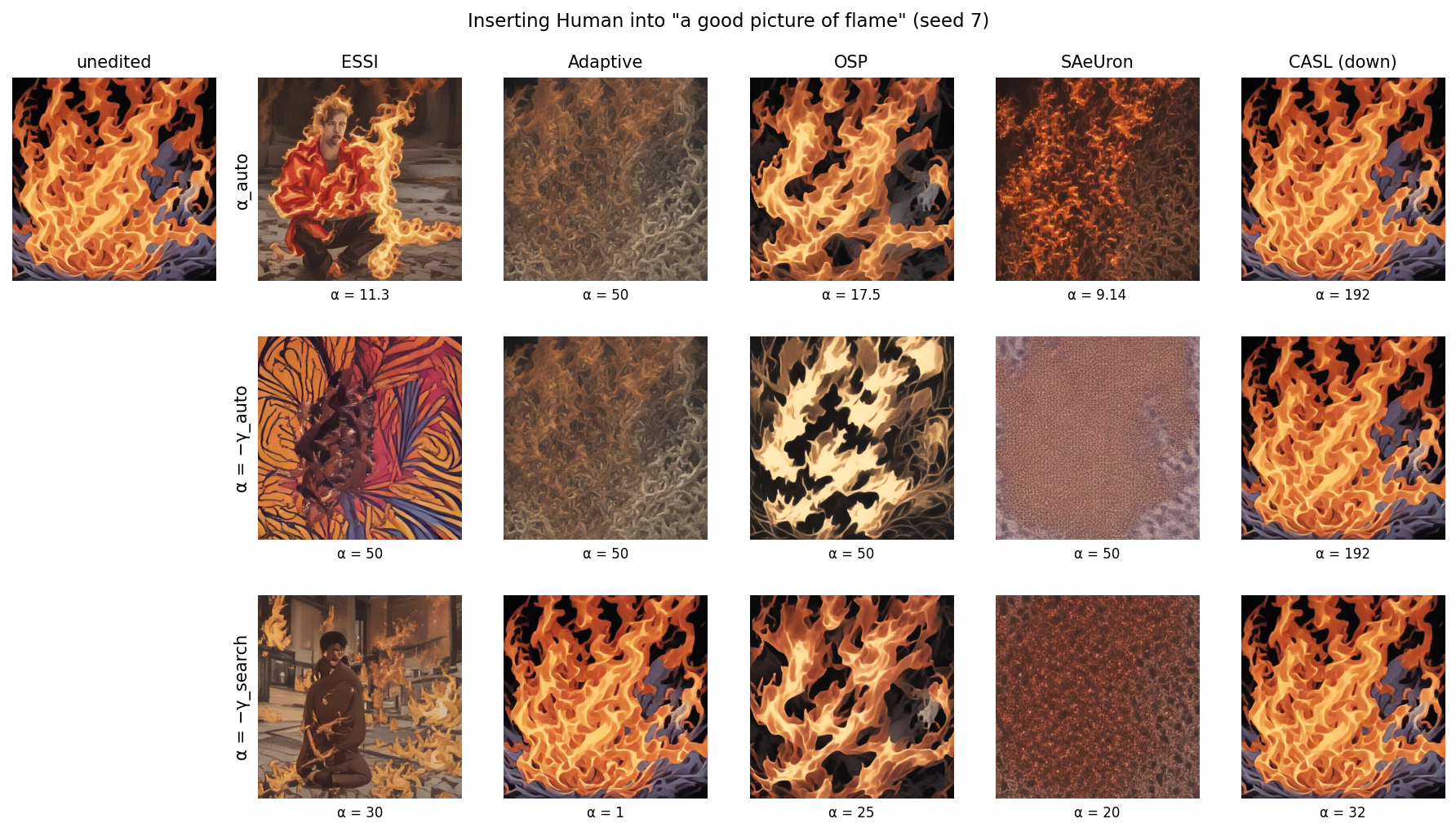}
    \includegraphics[scale=0.25]{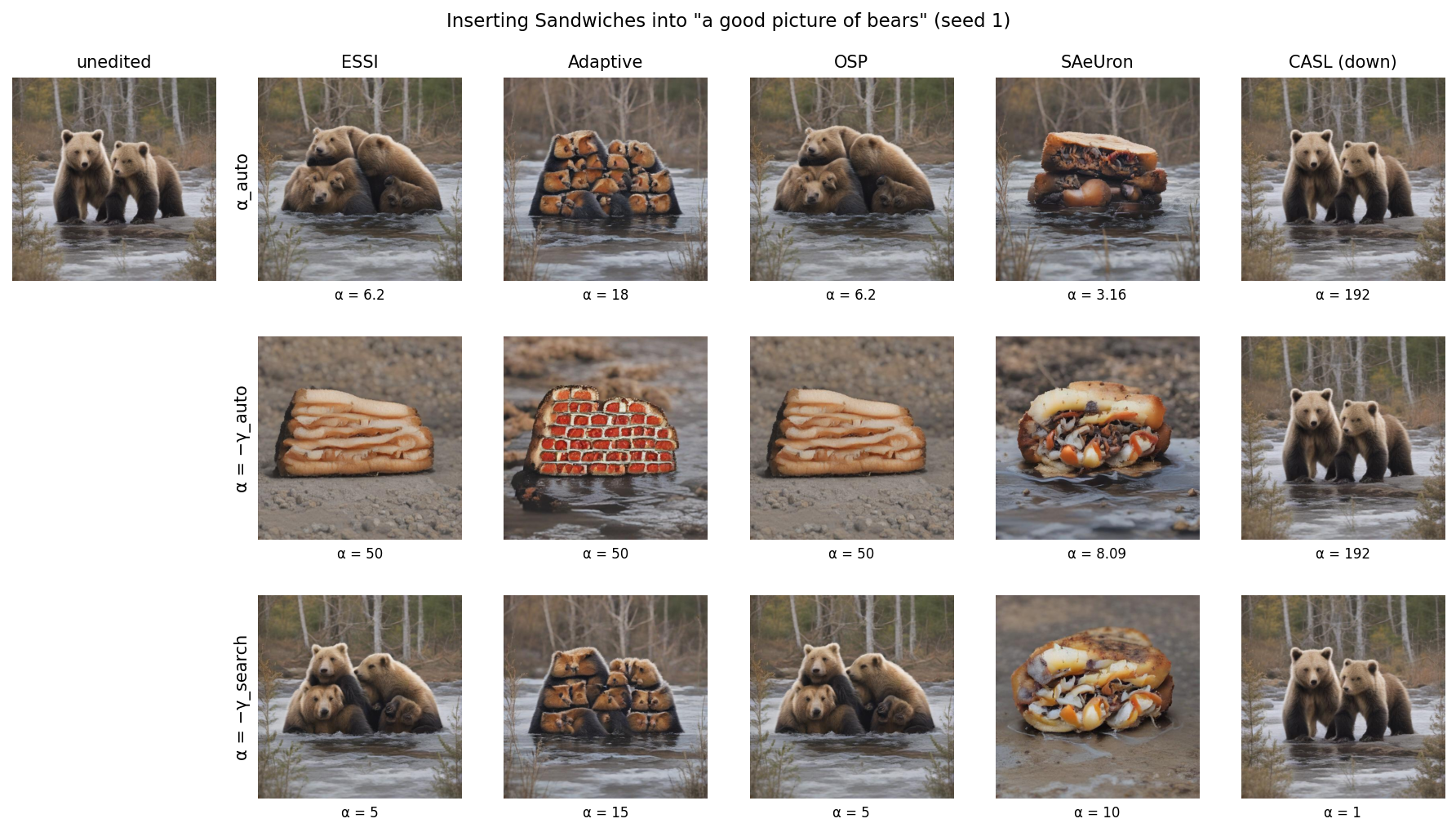}
    \caption{Concept insertion examples at \texttt{down.2.1}. Each panel shows the unedited base image (left) and each method's edit with \(\alpha_{\text{auto}}\), \(\alpha_{\gamma_{\text{auto}}}\) and \(\alpha_{\gamma_{\text{search}}}\) (rows, top to bottom). Top: inserting \emph{Human} into ``a good picture of flame.'' Bottom: inserting \emph{Sandwiches} into ``a good picture of bears.''}\label{fig:insertion}
\end{figure}

\paragraph{Evaluation images.}
The evaluation set is one image per (object, style) pair: the prompt ``A \(o\) image in \(s\) style.'' for every object \(o\) and each of the 50 UnlearnCanvas styles \(s\), generated with a fixed seed, so \(20 \times 50 = 1000\) prompts.
We write \(x_{o,s}\) for the unedited image of a prompt and \(\tilde{x}_{o,s}\) for the image generated from the same prompt and seed with an edit applied.
All experiments use a single seed; reported numbers are means over the 20 concepts. Removal and insertion examples are shown in Figs.~\ref{fig:removal} and~\ref{fig:insertion}.

\subsection{Removal}
\label{sec:exp:removal}
To remove a concept \(c\), we regenerate \emph{every} prompt of the evaluation set with the same seed while applying the edit of Section~\ref{sec:method:removal} for \(c\), and compare each edited image \(\tilde{x}_{o,s}\) with the unedited image \(x_{o,s}\).
The prompts with \(o = c\) (50 images) are the \emph{target} images; the other 950 measure collateral damage.
We report, averaged over the 20 concepts:
\begin{enumerate}
    \item \textbf{Unlearning accuracy (UA)}: the fraction of target images that are no longer classified as \(c\),
          \(\frac{1}{50}\sum_{s} \mathbb{1}[f_{\text{obj}}(\tilde{x}_{c,s}) \neq c]\). Higher is better.
    \item \textbf{In-domain retain accuracy (IRA)}: the object classifier's accuracy on the non-target images, i.e.\ whether the other objects survive the removal of \(c\),
          \(\frac{1}{950}\sum_{o \neq c,\, s} \mathbb{1}[f_{\text{obj}}(\tilde{x}_{o,s}) = o]\). Higher is better.
    \item \textbf{Cross-domain retain accuracy (CRA)}: the style classifier's accuracy on the same non-target images, i.e.\ whether removing an object leaves the styles intact,
          \(\frac{1}{950}\sum_{o \neq c,\, s} \mathbb{1}[f_{\text{sty}}(\tilde{x}_{o,s}) = s]\). Higher is better.
    \item \textbf{Fr\'echet Inception Distance (FID)} \citep{heusel2017gans}: the Fr\'echet distance between Gaussians fitted to the Inception-v3 pool features of the edited images and of the unedited images of the same 1000 prompts. It measures how far the edit moves the image distribution, not image realism against real data. Lower is better.
    \item \textbf{(UA+IRA)/2}: Each metric alone can be gamed: replacing the image with noise maximizes UA, and leaving it unchanged maximizes IRA. Their mean is high only when the target concept is removed and unrelated content is preserved.
\end{enumerate}
For every feature selection method, we compare two choices of the removal strength \(\gamma\): \(\gamma_{\text{auto}}\), found automatically as described in Section~\ref{sec:method:removal}, and \(\gamma_{\text{search}}\), found by grid search as in the codebase of SAeUron \citep{cywinski2025saeuroninterpretableconceptunlearning}.
For \(\gamma_{\text{search}}\), each \(\gamma \in \{-1, -5, -10, -15, -20, -25, -30\}\) is applied to SAeUron's sweep prompts, and the value maximizing (UA+IRA)/2 is kept; for SAeUron, the number of features \(\tau \in \{1, 2, 5, 10, 15, 20, 25, 30\}\) is searched jointly with \(\gamma\).
CASL's strength lives on a different scale, so for CASL we search its steering strength \(\alpha \in \{1, 32, 64, 96, 128\}\) instead and remove with \(\gamma = -\alpha\).
Following SAeUron's code, the search is scored with the released (not retrained) object classifier.
Its prompts use a different template from the evaluation prompts, but cover the same 50 styles, so \(\gamma_{\text{search}}\) is not tuned on fully held-out data; \(\gamma_{\text{auto}}\) uses no classifier and no evaluation prompts.
Results are in Table~\ref{table:removal}.

\begin{table*}[t]
\centering
\caption{Concept removal results at \texttt{down.2.1} (CASL also at \texttt{mid.0}), means over the 20 UnlearnCanvas objects. Best in \textbf{bold}, second best \textit{italicized}.}
\label{table:removal}
\begin{tabular}{|lccccc|}
\hline
    Method & UA \(\uparrow\) & IRA \(\uparrow\) & CRA \(\uparrow\) & FID \(\downarrow\) & (UA+IRA)/2 \(\uparrow\) \\
    \hline
    \multicolumn{6}{|c|}{\(\gamma_{\text{auto}}\)} \\
    \hline
    ESSI & 0.676 & \textit{0.990} & \textbf{0.980} & \textit{9.897} & \textit{0.833} \\
    Adaptive & 0.668 & \textbf{0.992} & \textbf{0.980} & \textbf{9.659} & 0.830 \\
    OSP & \textit{0.678} & 0.981 & \textit{0.973} & 11.439 & 0.830 \\
    SAeUron & \textbf{0.722} & 0.978 & 0.966 & 14.091 & \textbf{0.850} \\
    CASL (down) & 0.285 & 0.722 & 0.711 & 89.733 & 0.503 \\
    CASL (mid) & 0.455 & 0.518 & 0.497 & 171.545 & 0.486 \\
    \hline
    \multicolumn{6}{|c|}{\(\gamma_{\text{search}}\)} \\
    \hline
    ESSI & \textit{0.827} & \textit{0.995} & \textbf{0.985} & 10.431 & \textit{0.911} \\
    Adaptive & 0.694 & \textbf{0.996} & \textbf{0.985} & \textbf{8.902} & 0.845 \\
    OSP & 0.615 & 0.993 & \textit{0.982} & \textit{9.274} & 0.804 \\
    SAeUron & \textbf{0.937} & 0.983 & 0.969 & 15.094 & \textbf{0.960} \\
    CASL (down) & 0.191 & 0.812 & 0.799 & 48.472 & 0.502 \\
    CASL (mid) & 0.254 & 0.738 & 0.720 & 79.391 & 0.496 \\
\hline
\end{tabular}
\end{table*}

\subsection{Insertion}
\label{sec:exp:insertion}
To insert a concept \(c\), we use a separate set of base images: the prompt ``a good picture of \(b\)'' for each of the 20 UnlearnCanvas objects \(b\), generated with the same fixed seed and segmented with SAM~3 on \(b\).
We regenerate each base prompt with the same seed while adding concept \(c\) inside the base subject's mask, as described in Section~\ref{sec:method:insertion}, giving an edited image \(\tilde{x}_b\) for every base \(x_b\).
Bases whose subject is \(c\) itself are skipped, leaving 19 edited images per concept.
We report the \textbf{Editing Precision Ratio (EPR)} \citep{he2026caslconceptalignedsparselatents}, averaged over the 20 concepts: the mean absolute change of the object classifier's logit for \(c\) between \(x_b\) and \(\tilde{x}_b\), divided by the mean absolute change of its other 19 object logits. It is high when the edit adds \(c\) without disturbing the classifier's other concepts. Higher is better.
For every feature selection method, we compare three insertion strengths: \(\alpha_{\text{auto}}\), found automatically as described in Section~\ref{sec:method:insertion}, and the negated removal strengths \(\alpha_{\gamma_{\text{auto}}} = -\gamma_{\text{auto}}\) and \(\alpha_{\gamma_{\text{search}}} = -\gamma_{\text{search}}\) of Section~\ref{sec:exp:removal}.
Results are in Table~\ref{table:insertion}.

\begin{table*}[t]
    \centering
    \caption{Concept insertion results at \texttt{down.2.1} (CASL also at \texttt{mid.0}), means over the 20 UnlearnCanvas objects. Best in \textbf{bold}, second best \textit{italicized}.}
    \label{table:insertion}
    \begin{tabular}{|lc|}
    \hline
       Method & EPR \(\uparrow\) \\
       \hline
       \multicolumn{2}{|c|}{\(\alpha_{\text{auto}}\)} \\
       \hline
       ESSI & 1.994 \\
       Adaptive & \textit{2.481} \\
       OSP & 1.860 \\
       SAeUron & \textbf{2.592} \\
       CASL (down)* & 0.536 \\
       CASL (mid)* & 0.637 \\
       \hline
       \multicolumn{2}{|c|}{\(\alpha_{\gamma_{\text{auto}}}\)} \\
       \hline
       ESSI & \textit{1.570} \\
       Adaptive & \textbf{1.760} \\
       OSP & 1.411 \\
       SAeUron & 1.395 \\
       CASL (down)* & 0.536 \\
       CASL (mid)* & 0.637 \\
       \hline
       \multicolumn{2}{|c|}{\(\alpha_{\gamma_{\text{search}}}\)} \\
       \hline
       ESSI & \textbf{2.641} \\
       Adaptive & \textit{2.520} \\
       OSP & 1.882 \\
       SAeUron & 2.476 \\
       CASL (down) & 0.510 \\
       CASL (mid) & 0.625 \\
       \hline
    \end{tabular}
    \caption*{*For CASL, neither automatic search reaches its target within its range (\(|\gamma|, \alpha \le 192\)) for any concept, so both fall back to \(\alpha = 192\) and give the same edit.}
\end{table*}

\subsection{Efficiency}
\label{sec:exp:efficiency}
We count the floating-point operations (FLOPs) each method needs to select its features (Table~\ref{table:flopfeature}) and its removal and insertion strengths (Table~\ref{table:flopstrength}), per concept at one block.
GPU work (generation, classification, segmentation, SAE encoding, CASL training) is counted with PyTorch's FLOP counter; CPU work (probes) is counted analytically.
FLOPs are operation counts, independent of numeric precision (generation ran in fp16); the time and energy per FLOP depend on the hardware and precision.
We write G, T and P for \(10^{9}\), \(10^{12}\) and \(10^{15}\) FLOPs. Results are in Table~\ref{table:flopfeature} and Table~\ref{table:flopstrength}.

\begin{table*}[t]
    \centering
    \caption{FLOPs per concept for feature selection.}
    \label{table:flopfeature}
    \begin{tabular}{|lc|}
    \hline
       Method & FLOPs/concept \\
       \hline
        ESSI & 27.8\,T \\
        Adaptive & 27.8\,T \\
        OSP & 27.9\,T \\
        SAeUron* & 250\,P \\
        CASL & 56.3\,P \\
    \hline
    \end{tabular}
    \caption*{*Includes SAeUron's search over \(\gamma\), which it performs jointly with its feature search.}
\end{table*}

\begin{table*}[t]
    \centering
    \caption{FLOPs per concept for strength selection, on top of feature selection.}
    \label{table:flopstrength}
    \begin{tabular}{|lc|}
    \hline
    Strength & FLOPs/concept \\
    \hline
      \(\gamma_{\text{auto}}\) & 104\,G \\
      \(\alpha_{\text{auto}}\) & 104\,G \\
      \(\gamma_{\text{search}}\) (ESSI, Adaptive, OSP) & 30.6\,P \\
      \(\gamma_{\text{search}}\) (CASL) & 22.3\,P \\
      \(\gamma_{\text{search}}\) (SAeUron)* & 0 \\
    \hline
    \end{tabular}
    \caption*{*SAeUron searches for its features and removal strength jointly, so its search cost is already in Table~\ref{table:flopfeature}.}
\end{table*}

\subsection{Discussion}
\label{sec:exp:discussion}
The most significant result is the drastic difference in efficiency between the ESSI, OSP and Adaptive methods, and the rest. ESSI requires roughly 9000 times fewer FLOPs than SAeUron, and roughly 2000 times fewer FLOPs than CASL. For removal, for both methods of choosing \(\gamma\), SAeUron still performs best on (UA+IRA)/2, while ESSI ranks second. Given that, it's more accurate to say that ESSI provides comparable removal for a fraction of the compute, albeit not superior. For insertion, the adaptive method is first or second best for all ways of choosing \(\alpha\), but note that ESSI with \(\alpha_{\gamma_{\text{search}}}\) is the highest score across all combinations of method and strength. ESSI + \(\alpha_{\gamma_{\text{search}}}\) would be cheaper than SAeUron or CASL and do extremely well at insertion and fairly well at removal. CASL performs poorly on all metrics. Our experiments reflected a fairly wide domain shift. The original paper used 50 inference steps on Stable Diffusion 1.5. Despite a thorough search over hyperparameters (including inference steps) in Appendix~\ref{appendix:impl:casl}, none of our CASL configurations performed well on our SDXL-Turbo model. CASL was also evaluated on facial features, while our method had a much more diverse set of concepts to learn.

\section{Conclusion}
\label{sec:conclusion}

This is the first study to evaluate the same selected features on both removal and insertion, comparing five methods. Our contribution, ESSI, turns a named concept into a small set of SAE features plus a removal strength and an insertion strength. It does this using segmentation masks and linear probes on cached activations, and generates no images during the search. We caveat that the ESSI, OSP and Adaptive methods tend to have very similar results, and almost identical efficiencies, but ESSI has a slight but consistent edge in removal compared to the other inexpensive methods. CASL is not robust to our choice of model and concepts, so further work would entail trying this analysis again on a larger set of models and subjects, so as to discover which techniques work on which configurations, and why. Another next step is to find a way to decide between layers, as it's possible that some of our target concepts were not present in all layers. Our work also assumes that sparse autoencoders are the best way to represent features; it is possible that features are not as polysemantic as people assume and we may not need them.

\subsubsection*{Author Contributions}
Redacted while under review

\subsubsection*{Acknowledgments}
Redacted while under review
{
    \small
    \bibliographystyle{tmlr}
    \bibliography{main}
}

 \section{Appendix}
\label{sec:supplementary}

\section{Implementation details}\label{appendix:implementation}
Every method edits SDXL-Turbo with a single denoising step and the same pretrained TopK SAEs \citep{surkov2024unpackingsdxlturbointerpreting} as ESSI, and reads the same cached activations of the concept images of Section~\ref{sec:method:activations} (25 prompts per concept, one seed).
Unless stated otherwise, removal and insertion are applied exactly as in Sections~\ref{sec:method:removal} and~\ref{sec:method:insertion}: removal rescales the selected features by \(\gamma_c\,\bar{\mu}_i\) and subtracts only their change through the SAE decoder, and insertion adds \(\alpha_c\, W^{\text{dec}}_l v\) inside the base subject's mask.
The strengths \(\gamma_{\text{auto}}\) and \(\alpha_{\text{auto}}\) are found with the same dense probes, targets and bisection as ESSI's, applying each method's own edit offline; \(\gamma_{\text{search}}\) uses the grid search of Section~\ref{sec:exp:removal}.
The methods therefore differ only in how they choose the features, except CASL, which also edits the activations in its own way.

\subsection{SAeUron}
\label{appendix:impl:saeuron}
We reimplement the feature selection and removal of SAeUron \citep{cywinski2025saeuroninterpretableconceptunlearning} on our SAEs, following its released code.
Each image's SAE code is averaged over \emph{all} of its patches, and features are scored with the score of Section~\ref{sec:method:score} computed on these unmasked means; the top \(\tau\) features are kept.
SAeUron does not choose \(\tau\) from data: the paper tunes \(\tau\) and \(\gamma\) per concept on a validation set, and the values it reports were tuned for an SAE trained on Stable Diffusion~1.5, so we rerun its search on our model.
For each concept, we evaluate every \(\tau \in \{1, 2, 5, 10, 15, 20, 25, 30\}\) jointly with every \(\gamma \in \{-1, -5, -10, -15, -20, -25, -30\}\), separately at each block, and keep the pair that maximizes (UA+IRA)/2.
The search images are SAeUron's sweep prompts, the UnlearnCanvas anchor prompt of each object followed by ``in \(s\) style.'' for 49 styles (\(20 \times 49 = 980\) prompts), generated with a fixed seed per prompt and scored with the released UnlearnCanvas object classifier, as in SAeUron's code.
Its released code thresholds the scores at fixed percentiles (99.99, 99.995, 99.999) rather than taking the top \(\tau\); with our 5120-feature SAEs every percentile keeps only the top feature, so we search \(\tau\) directly, as described in the paper.
Removal follows SAeUron's masked edit: feature \(i \in I\) is rescaled to \(\gamma\,\bar{\mu}_i\, z_i\) only at patches where \(z_i\) exceeds its mean activation over all concepts (the mean over concepts of each concept's mean image code), and left unchanged elsewhere.
For \(\gamma_{\text{auto}}\), we keep the searched \(\tau\) and replace only \(\gamma\) by the value of the probe bisection, applying the thresholded edit offline.
Insertion adds the searched features exactly as ESSI adds its own (Section~\ref{sec:method:insertion}).

\subsection{CASL}
\label{appendix:impl:casl}
CASL \citep{he2026caslconceptalignedsparselatents} learns, for each concept, a linear map from the frozen SAE code \(z\) of a block to an edit of the block's output, \(\Delta h = W_\Delta z + b_\Delta\) with \(W_\Delta \in \mathbb{R}^{D_l \times F}\), and selects the features with the largest columns of \(W_\Delta\).
We train it as the paper's second stage.
The map is added at every patch of the block (\texttt{down.2.1} or \texttt{mid.0}), with \(z\) the code of the block's residual update, as for the other methods.
The training loss is DiffusionCLIP's directional loss plus an \(\ell_1\) reconstruction term,
\begin{equation}
    \mathcal{L} = \lambda_{\text{CLIP}} \Bigl(1 - \cos\bigl(E_I(\tilde{x}) - E_I(x),\; E_T(y_{\text{ref}}) - E_T(y_{\text{origin}})\bigr)\Bigr) + \lambda_{\text{rec}}\, \lVert \tilde{x} - x \rVert_1,
\end{equation}
where \(x\) is the unedited image, \(\tilde{x}\) the image generated from the same prompt and seed with the map applied, \(E_I, E_T\) the CLIP ViT-B/16 image and text encoders, \(\lambda_{\text{CLIP}} = 3\) and \(\lambda_{\text{rec}} = 1\).
The text pair goes from the domain to the concept, \(y_{\text{origin}} = \)``a photo'' and \(y_{\text{ref}} = \)``a photo of a \(c\)''.
Because SDXL-Turbo produces the image in one step, \(x\) is generated directly and no DDIM inversion is needed; gradients flow through the U-Net and the VAE decoder to \(W_\Delta\) and \(b_\Delta\) only.
The paper trains on 1,000 images per concept without specifying them; we use one pool shared by all concepts, the concept prompts of all 20 objects with two seeds each (\(20 \times 25 \times 2 = 1000\) images).
The paper does not give an optimizer, learning rate or number of steps.
We initialize \(W_\Delta\) and \(b_\Delta\) at zero, so that features that never fire on the training images keep a zero column and do not enter the ranking, and train with Adam (learning rate \(10^{-2}\), one image per step) for 5,000 steps; in a sweep over optimizers and learning rates, this was the setting whose CLIP loss clearly decreased the most. Refer to the Table \ref{tab:hyper} for an exhaustive search of hyperparameters.

\begin{table}[]
\begin{tabular}{|cccccc|}
\hline
block & training steps & lr & inference steps & optimizer & \(\Delta\) \(\mathcal{L}_{CLIP}\) \\
\hline
down.2.1 & 1000 & 1e-5 & 1 & SGD & -0.003 \\
down.2.1 & 1000 & 1e-4 & 1 & SGD & -0.005 \\
down.2.1 & 1000 & 1e-3 & 1 & Adam & 0.024 \\
down.2.1 & 1000 & 1e-3 & 1 & SGD & 0.017 \\
down.2.1 & 1000 & 1e-3 & 2 & Adam & -0.011 \\
down.2.1 & 1000 & 1e-3 & 10 & Adam & 0.012 \\
down.2.1 & 1000 & 1e-2 & 1 & Adam & 0.054 \\
down.2.1 & 1000 & 1e-2 & 2 & Adam & 0.013 \\
down.2.1 & 1000 & 1e-2 & 10 & Adam & -0.002 \\
down.2.1 & 3000 & 1e-3 & 1 & Adam & 0.043 \\
down.2.1 & 3000 & 1e-3 & 2 & Adam & 0.002 \\
down.2.1 & 3000 & 1e-3 & 10 & Adam & 0.003 \\
down.2.1 & 3000 & 1e-2 & 1 & Adam & 0.125 \\
down.2.1 & 3000 & 1e-2 & 2 & Adam & -0.001 \\
down.2.1 & 3000 & 1e-2 & 10 & Adam & 0.003 \\
down.2.1 & 5000 & 1e-2 & 1 & Adam & 0.138 \\
down.2.1 & 10000 & 1e-2 & 1 & Adam & 0.138 \\
mid & 1000 & 1e-5 & 1 & SGD & -0.008 \\
mid & 1000 & 1e-4 & 1 & SGD & 0.003 \\
mid & 1000 & 1e-3 & 1 & Adam & 0.024 \\
mid & 1000 & 1e-3 & 1 & SGD & -0.009 \\
mid & 1000 & 1e-3 & 2 & Adam & 0.013 \\
mid & 1000 & 1e-3 & 10 & Adam & -0.024 \\
mid & 1000 & 1e-2 & 1 & Adam & 0.092 \\
mid & 1000 & 1e-2 & 2 & Adam & 0.005 \\
mid & 1000 & 1e-2 & 10 & Adam & 0.030 \\
mid & 3000 & 1e-3 & 1 & Adam & 0.049 \\
mid & 3000 & 1e-3 & 2 & Adam & 0.006 \\
mid & 3000 & 1e-3 & 10 & Adam & 0.017 \\
mid & 3000 & 1e-2 & 1 & Adam & 0.157 \\
mid & 3000 & 1e-2 & 2 & Adam & -0.013 \\
mid & 3000 & 1e-2 & 10 & Adam & -0.012 \\
mid & 5000 & 1e-2 & 1 & Adam & \textbf{0.171} \\
mid & 10000 & 1e-2 & 1 & Adam & 0.144 \\
\hline
\end{tabular}

    \caption{N=2 Hyperparameter search}
    \label{tab:hyper}
\end{table}

Features are ranked by the \(\ell_2\) norm of their column of \(W_\Delta\), and, as in the paper, only the top feature \(j\) is used.

CASL edits the block with its own map rather than through the SAE decoder (CASL-Steer):
\begin{equation}
    h' = h + \alpha\, W_\Delta[:, j]\, z_j
\end{equation}
at every patch, without the bias \(b_\Delta\) and without a mask, for both removal and insertion.
Since the map is trained to add the concept, insertion uses \(\alpha > 0\) and removal \(\alpha = -\gamma > 0\) in the opposite direction, \(h' = h + \gamma\, W_\Delta[:, j]\, z_j\).
The edit is proportional to \(z_j\), so it is zero at patches where the feature is inactive.
Because \(W_\Delta\) and the SAE decoder have different scales, CASL's strengths lie on a different scale from the other methods': the grid search uses \(\alpha \in \{1, 32, 64, 96, 128\}\), and the probe bisections search \(|\gamma|, \alpha \le 192\), the largest strength shown in the paper.

\subsection{OSP}
\label{appendix:impl:osp}
Optimal sparse probing (OSP) \citep{gurnee2023findingneuronshaystackcase} finds, for each \(k\), the \(k\) features of the best \(k\)-sparse linear classifier of the concept.
We follow the official implementation and its defaults.
The probe separates the same patches as ESSI's probes, the positive patches \(\mathcal{D}^{+}_{l,c}\) inside the concept's masks from the negative patches \(\mathcal{D}^{-}_{l,c}\) outside them, over all of the concept's images.
Each feature is divided by 10 times its standard deviation over these patches, and the 50 features with the largest absolute difference between the positive and negative means are kept as candidates.
For each \(k\), OSP solves the \(k\)-sparse hinge-loss SVM
\begin{equation}
    \min_{w, b} \; \sum_n c_n \max\bigl(0,\, 1 - y_n (w^\top z_n + b)\bigr) + \frac{1}{2\rho} \lVert w \rVert_2^2
    \quad \text{s.t.} \quad \lVert w \rVert_0 \le k
\end{equation}
over the candidates, with balanced class weights \(c_n\) and \(\rho = 1/\sqrt{N}\) for \(N\) patches, by the outer-approximation (cutting-plane) method of \citet{Bertsimas_2021_osp}: an SVM solved on a support gives a cut valid for every support, and a mixed-integer program over the cuts proposes the next support until the bound meets the best objective.
The official code solves both problems with Gurobi; we solve the SVM with libsvm and the mixed-integer program with HiGHS.
As in the official code, each solve is limited to 60\,s, after which the best support found is used, and \(k\) runs over \(8, 7, \dots, 1\), each solve warm-started from the top \(|w|\) of an \(\ell_2\) logistic regression on the previous, larger support.
The supports for different \(k\) need not be nested, so the number of features is chosen as in Section~\ref{sec:method:k} but among these eight supports: the smallest \(k\) whose support reaches 95\% of the held-out accuracy of the probe on all features, found by binary search over \(k \in \{1, \dots, 8\}\).

\subsection{Adaptive thresholding}
\label{appendix:impl:adaptive}
Adaptive thresholding (Adaptive) \citep{gurnee2023findingneuronshaystackcase} prunes the features iteratively with elastic-net logistic regressions.
It uses the same patches, normalization and mean-difference ranking as OSP, again following the official implementation.
Starting from the 64 features with the largest mean difference, it fits, for each \(k\) in the official schedule \(64, 32, 16, 12, 10, 8, 7, \dots, 1\) in turn, an elastic-net logistic regression with balanced class weights, \(C = k/20\) and \(\ell_1\) ratio \(\text{clip}(e^{-k/20}, 0.1, 0.9)\) on the current \(k\) features, and keeps the \(k'\) features with the largest absolute coefficients for the next, smaller \(k'\).
The resulting sets are nested, so they define a single ranking: the features of the smallest set come first, followed by the features each larger set adds, ordered by the absolute coefficients of that set's probe.
The number of features is then chosen from this ranking exactly as ESSI's (Section~\ref{sec:method:k}), by binary search over \(k \le 64\).

\subsection{Scores By Subject}
Some subjects were "harder" than others. For any concept, there is no guarantee that the SDXL-Turbo model was capable of representing it with a few sparse latents. Insertion, Removal FID and Removal (UA+IRA)/2 are in Table \ref{table:insertion_epr_per_object}, \ref{table:removal_fid_per_object} and \ref{table:removal_ua_ira_per_object}, respectively.

\begin{table*}[t]
\centering
\caption{Per-object insertion EPR \(\uparrow\) at \texttt{down.2.1}. In each column, the best object is in \textbf{bold}, the second best \textit{italicized} and the worst \underline{underlined}; the last row is the EPR pooled over all edited images (Table~\ref{table:insertion}).}
\label{table:insertion_epr_per_object}
\resizebox{\textwidth}{!}{%
\begin{tabular}{|l|ccccc|ccccc|ccccc|}
\hline
    & \multicolumn{5}{c|}{\(\alpha_{\text{auto}}\)} & \multicolumn{5}{c|}{\(\alpha_{\gamma_{\text{auto}}}\)} & \multicolumn{5}{c|}{\(\alpha_{\gamma_{\text{search}}}\)} \\
    Object & ESSI & Adaptive & OSP & SAeUron & CASL (down) & ESSI & Adaptive & OSP & SAeUron & CASL (down) & ESSI & Adaptive & OSP & SAeUron & CASL (down) \\
\hline
    Architectures & 1.085 & \textbf{5.784} & 2.888 & 1.339 & \underline{0.000} & 2.585 & 3.194 & 1.449 & \textit{2.436} & \underline{0.000} & 2.379 & 1.149 & 2.300 & \underline{0.518} & \underline{0.000} \\
    Bears & 1.464 & 0.907 & 0.907 & 1.035 & 0.395 & 0.531 & 0.871 & 0.871 & 0.683 & 0.395 & 1.636 & 1.071 & 1.071 & 1.032 & 0.464 \\
    Birds & 3.267 & 3.267 & 1.029 & 3.164 & \underline{0.000} & \underline{0.000} & \underline{0.000} & 0.549 & \underline{0.000} & \underline{0.000} & 3.579 & 3.579 & 1.050 & 3.285 & \underline{0.000} \\
    Butterfly & 1.494 & 0.762 & 0.762 & 1.787 & \underline{0.000} & 0.722 & 0.935 & 0.935 & 0.596 & \underline{0.000} & 1.582 & 0.865 & 0.865 & 1.201 & \underline{0.000} \\
    Cats & \textit{5.290} & 5.290 & \textbf{5.290} & \textbf{5.495} & \underline{0.000} & 1.279 & 1.279 & 1.279 & 0.999 & \underline{0.000} & 4.132 & \textit{4.132} & \textit{4.132} & 1.796 & \underline{0.000} \\
    Dogs & \textbf{5.389} & \textit{5.389} & 0.987 & \textit{4.587} & \underline{0.000} & 1.855 & 1.855 & 0.628 & 1.234 & \underline{0.000} & \textbf{5.348} & \textbf{5.348} & 1.078 & 3.977 & \underline{0.000} \\
    Fishes & 2.240 & 2.240 & 2.240 & 1.018 & \textit{0.582} & 0.838 & 0.838 & 0.838 & 0.529 & \textit{0.582} & 0.807 & 0.807 & 0.807 & 0.573 & \textit{0.668} \\
    Flame & 2.051 & 2.051 & 1.313 & 1.873 & \underline{0.000} & \textbf{3.493} & \textbf{3.493} & 2.311 & 1.074 & \underline{0.000} & 3.197 & 3.197 & 0.745 & 3.768 & \underline{0.000} \\
    Flowers & 0.717 & 4.348 & \textit{4.348} & 3.133 & \textbf{0.915} & 0.717 & 2.813 & \textit{2.813} & 1.360 & \textbf{0.915} & 1.264 & 3.953 & 3.953 & 2.483 & \textbf{1.115} \\
    Frogs & 0.820 & 0.820 & 0.820 & 0.820 & 0.511 & 0.738 & 0.738 & 0.738 & 0.738 & 0.511 & 1.001 & 1.001 & 1.001 & 1.001 & 0.591 \\
    Horses & 0.653 & 0.653 & 2.356 & 3.181 & \underline{0.000} & 0.653 & 0.653 & 0.585 & 1.126 & \underline{0.000} & 3.838 & 3.838 & 1.280 & 3.158 & \underline{0.000} \\
    Human & 3.017 & 1.195 & 0.995 & 1.161 & \underline{0.000} & 1.999 & 1.195 & 0.817 & 0.938 & \underline{0.000} & 2.601 & 1.058 & 0.842 & 1.177 & \underline{0.000} \\
    Jellyfish & 0.809 & 0.771 & 0.771 & 0.967 & \underline{0.000} & 0.809 & 0.529 & \underline{0.529} & 0.767 & \underline{0.000} & \underline{0.613} & \underline{0.653} & \underline{0.653} & 1.274 & \underline{0.000} \\
    Rabbits & \underline{0.467} & \underline{0.467} & \underline{0.467} & \underline{0.467} & 0.378 & 0.794 & 0.794 & 0.794 & 0.794 & 0.378 & 0.911 & 0.911 & 0.911 & 0.911 & 0.434 \\
    Sandwiches & 0.870 & 1.284 & 0.870 & 2.500 & \underline{0.000} & 1.445 & 1.277 & 1.445 & 1.365 & \underline{0.000} & 0.897 & 1.306 & 0.897 & 1.258 & \underline{0.000} \\
    Sea & 2.082 & 1.450 & 1.450 & 4.397 & \underline{0.000} & \underline{0.000} & 2.150 & 2.150 & 1.486 & \underline{0.000} & 3.539 & 0.894 & 0.894 & \textit{4.799} & \underline{0.000} \\
    Statues & 1.997 & 3.629 & 1.902 & 1.579 & 0.520 & 1.274 & 1.234 & 1.226 & 1.233 & 0.520 & \textit{4.568} & 3.254 & \textbf{4.213} & \textbf{5.132} & 0.516 \\
    Towers & 1.529 & 2.281 & 2.281 & 4.506 & \underline{0.000} & 2.048 & \textit{3.422} & \textbf{3.422} & \textbf{3.036} & \underline{0.000} & 2.797 & 1.189 & 1.189 & 4.754 & \underline{0.000} \\
    Trees & 1.271 & 1.026 & 1.026 & 2.984 & \underline{0.000} & \textit{3.045} & 1.443 & 1.443 & 1.513 & \underline{0.000} & 2.331 & 1.138 & 1.138 & 2.121 & \underline{0.000} \\
    Waterfalls & 3.344 & 3.344 & 3.344 & 3.344 & 0.434 & 1.707 & 1.707 & 1.707 & 1.707 & 0.434 & 2.703 & 2.703 & 2.703 & 2.703 & 0.342 \\
\hline
    Pooled & 1.994 & 2.481 & 1.860 & 2.592 & 0.536 & 1.570 & 1.760 & 1.411 & 1.395 & 0.536 & 2.641 & 2.520 & 1.882 & 2.476 & 0.510 \\
\hline
\end{tabular}}
\end{table*}

\begin{table*}[t]
\centering
\caption{Per-object removal FID \(\downarrow\) at \texttt{down.2.1}. In each column, the best object is in \textbf{bold}, the second best \textit{italicized} and the worst \underline{underlined}; the last row is the mean over the 20 objects (Table~\ref{table:removal}).}
\label{table:removal_fid_per_object}
\resizebox{\textwidth}{!}{%
\begin{tabular}{|l|ccccc|ccccc|}
\hline
    & \multicolumn{5}{c|}{\(\gamma_{\text{auto}}\)} & \multicolumn{5}{c|}{\(\gamma_{\text{search}}\)} \\
    Object & ESSI & Adaptive & OSP & SAeUron & CASL (down) & ESSI & Adaptive & OSP & SAeUron & CASL (down) \\
\hline
    Architectures & 3.226 & 7.262 & 15.499 & 16.054 & \textbf{0.000} & 3.052 & 1.539 & 13.324 & \underline{44.670} & \textbf{0.000} \\
    Bears & 14.105 & 6.244 & 6.244 & 11.176 & 237.543 & 8.833 & 4.815 & 4.815 & 8.003 & 130.334 \\
    Birds & \textit{0.881} & \textit{0.881} & \underline{25.082} & \textbf{0.884} & 0.223 & 9.801 & 9.801 & \underline{16.665} & 8.673 & 0.015 \\
    Butterfly & 9.823 & 10.847 & 10.847 & 6.775 & \textbf{0.000} & 10.712 & 11.204 & 11.204 & 11.860 & \textbf{0.000} \\
    Cats & 5.321 & 5.321 & 5.321 & \textit{4.322} & \textbf{0.000} & 13.554 & 13.554 & 13.554 & \textit{7.938} & \textbf{0.000} \\
    Dogs & 12.243 & 12.243 & 19.842 & 5.587 & 0.736 & 7.575 & 7.575 & 2.170 & 11.197 & 0.073 \\
    Fishes & 15.054 & 15.054 & 15.054 & 33.296 & 215.814 & 14.923 & 14.923 & 14.923 & 23.867 & 110.904 \\
    Flame & 15.285 & 15.285 & 9.552 & 4.414 & 4.321 & 13.058 & 13.058 & \textit{1.319} & 12.864 & 0.201 \\
    Flowers & 4.673 & 20.070 & 20.049 & 21.582 & \underline{373.302} & \textit{3.020} & 14.570 & 14.570 & 15.674 & 9.415 \\
    Frogs & 8.419 & 8.419 & 8.419 & 8.419 & 226.734 & 6.139 & 6.139 & 6.139 & \textbf{6.143} & 123.673 \\
    Horses & 20.314 & \underline{20.314} & 12.022 & 9.829 & \textit{0.093} & 18.213 & \underline{18.213} & 9.192 & 20.017 & \textit{0.003} \\
    Human & \textbf{0.523} & 6.709 & 17.498 & 14.434 & 0.253 & \textbf{0.476} & \textit{1.304} & 14.062 & 9.972 & 0.301 \\
    Jellyfish & 10.536 & 7.969 & 7.969 & 12.528 & 2.588 & 8.801 & 6.974 & 6.974 & 8.870 & 0.114 \\
    Rabbits & 5.300 & 5.300 & \textit{5.281} & 5.281 & 224.723 & 9.011 & 9.011 & 9.011 & 9.007 & 140.859 \\
    Sandwiches & 15.876 & 16.982 & 15.876 & 16.089 & \textbf{0.000} & 11.909 & 13.266 & 11.909 & 16.404 & \textbf{0.000} \\
    Sea & 1.232 & \textbf{0.353} & \textbf{0.353} & 4.408 & 1.421 & 9.509 & \textbf{0.061} & \textbf{0.061} & 10.431 & 0.042 \\
    Statues & 7.273 & 7.351 & 7.328 & 6.620 & 283.528 & 13.204 & 10.939 & 14.496 & 11.693 & \underline{268.301} \\
    Towers & 10.985 & 6.986 & 6.986 & \underline{50.164} & \textbf{0.000} & 11.538 & 2.000 & 2.000 & 20.200 & \textbf{0.000} \\
    Trees & \underline{25.013} & 7.744 & 7.744 & 37.942 & 0.123 & \underline{21.288} & 5.074 & 5.074 & 30.381 & 0.108 \\
    Waterfalls & 11.862 & 11.862 & 11.862 & 11.862 & 223.258 & 14.009 & 14.009 & 14.009 & 14.016 & 185.091 \\
\hline
    Mean & 9.897 & 9.660 & 11.441 & 14.083 & 89.733 & 10.431 & 8.902 & 9.274 & 15.094 & 48.472 \\
\hline
\end{tabular}}
\end{table*}

\begin{table*}[t]
\centering
\caption{Per-object removal (UA+IRA)/2 \(\uparrow\) at \texttt{down.2.1}. In each column, the best object is in \textbf{bold}, the second best \textit{italicized} and the worst \underline{underlined}; the last row is the mean over the 20 objects (Table~\ref{table:removal}).}
\label{table:removal_ua_ira_per_object}
\resizebox{\textwidth}{!}{%
\begin{tabular}{|l|ccccc|ccccc|}
\hline
    & \multicolumn{5}{c|}{\(\gamma_{\text{auto}}\)} & \multicolumn{5}{c|}{\(\gamma_{\text{search}}\)} \\
    Object & ESSI & Adaptive & OSP & SAeUron & CASL (down) & ESSI & Adaptive & OSP & SAeUron & CASL (down) \\
\hline
    Architectures & 0.689 & 0.609 & 0.918 & 0.629 & 0.500 & 0.640 & \underline{0.500} & 0.854 & 0.865 & 0.500 \\
    Bears & 0.990 & 0.766 & 0.766 & 0.993 & \textbf{0.572} & \textbf{1.000} & 0.679 & 0.679 & \textbf{1.000} & \textbf{0.643} \\
    Birds & 0.510 & 0.510 & 0.539 & \underline{0.510} & 0.499 & 0.995 & 0.995 & \underline{0.483} & 0.996 & 0.500 \\
    Butterfly & \textbf{1.000} & \textit{0.999} & \textit{0.999} & 0.950 & 0.500 & \textbf{1.000} & \textbf{1.000} & \textbf{1.000} & \textbf{1.000} & 0.500 \\
    Cats & 0.790 & 0.790 & 0.790 & 0.710 & 0.500 & 0.995 & 0.995 & 0.995 & \textbf{1.000} & 0.500 \\
    Dogs & \textit{0.999} & \textit{0.999} & 0.624 & 0.730 & 0.497 & 0.990 & 0.990 & 0.500 & \textit{0.999} & 0.500 \\
    Fishes & 0.995 & 0.995 & 0.995 & 0.933 & 0.476 & 0.998 & 0.998 & 0.998 & 0.962 & \underline{0.383} \\
    Flame & 0.548 & 0.548 & 0.510 & 0.520 & 0.487 & 0.831 & 0.831 & 0.500 & \underline{0.691} & 0.500 \\
    Flowers & 0.586 & 0.966 & 0.966 & 0.962 & 0.527 & 0.559 & 0.966 & 0.966 & 0.965 & 0.500 \\
    Frogs & \textbf{1.000} & \textbf{1.000} & \textbf{1.000} & \textbf{1.000} & 0.523 & \textbf{1.000} & \textbf{1.000} & \textbf{1.000} & \textbf{1.000} & 0.479 \\
    Horses & 0.987 & 0.987 & \textbf{1.000} & 0.990 & 0.499 & 0.995 & 0.995 & \textbf{1.000} & 0.996 & 0.500 \\
    Human & 0.510 & 0.645 & 0.741 & 0.913 & 0.500 & \underline{0.510} & \underline{0.500} & 0.655 & 0.869 & 0.500 \\
    Jellyfish & \textbf{1.000} & 0.930 & 0.930 & 0.998 & 0.493 & 0.930 & 0.870 & 0.870 & 0.969 & 0.500 \\
    Rabbits & 0.760 & 0.760 & 0.760 & 0.760 & 0.517 & \textit{0.999} & \textit{0.999} & \textit{0.999} & \textit{0.999} & 0.526 \\
    Sandwiches & 0.984 & 0.987 & 0.984 & \textit{0.999} & 0.500 & \textbf{1.000} & \textit{0.999} & \textbf{1.000} & \textit{0.999} & 0.500 \\
    Sea & \underline{0.500} & \underline{0.500} & \underline{0.500} & 0.560 & 0.495 & 0.880 & \underline{0.500} & 0.500 & 0.940 & 0.500 \\
    Statues & 0.920 & 0.940 & 0.900 & 0.930 & \textit{0.554} & \textbf{1.000} & \textbf{1.000} & \textbf{1.000} & \textbf{1.000} & \textit{0.572} \\
    Towers & 0.929 & 0.950 & 0.950 & 0.940 & 0.500 & 0.920 & 0.550 & 0.550 & 0.981 & 0.500 \\
    Trees & 0.965 & 0.719 & 0.719 & 0.949 & 0.499 & 0.981 & 0.530 & 0.530 & 0.969 & 0.499 \\
    Waterfalls & \textbf{1.000} & \textbf{1.000} & \textbf{1.000} & \textbf{1.000} & \underline{0.428} & \textbf{1.000} & \textbf{1.000} & \textbf{1.000} & \textbf{1.000} & 0.429 \\
\hline
    Mean & 0.833 & 0.830 & 0.830 & 0.849 & 0.503 & 0.911 & 0.845 & 0.804 & 0.960 & 0.502 \\
\hline
\end{tabular}}
\end{table*}

\subsection{Classifier Comparison}\label{appendix:cra}
The classifier used in the original UnlearnCanvas repository was fine-tuned on images generated by Stable Diffusion 1.5, which is a different model than what we used (SDXL-Turbo). This domain shift was surprisingly impactful and their style classifier performed far worse on style classification, and slightly worse on object classification on a random sample of 100 images generated by our SDXL-Turbo model, as shown in Table \ref{tab:class}. We replicated the training of the object and style classifiers using the same methodology to train the original classifiers, with the difference that we used images generated by SDXL-Turbo.

\begin{figure}
    \centering
    \includegraphics[scale=0.5]{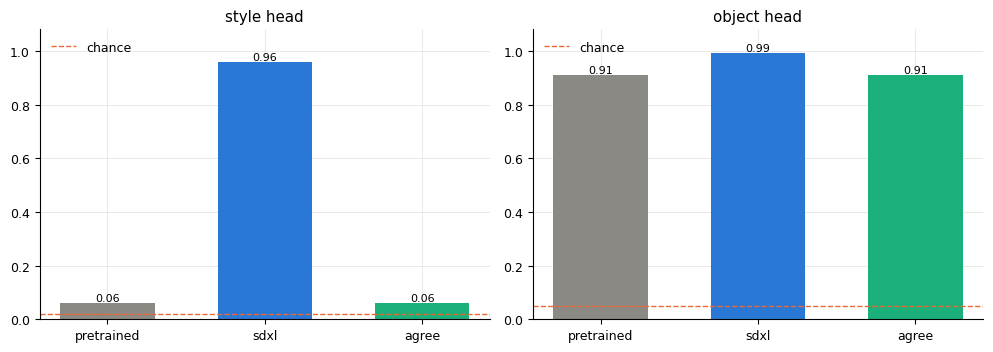}
    \caption{SDXL Classifier vs. Pretrained Classifier Visualization}
\end{figure}

\begin{table*}
\centering
\caption{SDXL Classifier vs. Pretrained Classifier Results}\label{tab:class}
    \begin{tabular}{|c|c|c|}
    \hline
        metric & style &  object\\
        \hline
        accuracy (pretrained) & 0.06 & 0.91 \\
        accuracy (sdxl) & 0.96 & 0.99 \\
        both right & 0.06 & 0.91 \\
        both wrong & 0.04 & 0.01 \\
        \hline
    \end{tabular}
\end{table*}

\subsection{Block Comparisons}\label{appendix:block}
A few preliminary experiments showed that the \texttt{down.2.1} block was better at both insertion and removal than the other blocks which we could find pretrained Sparse Autoencoders for. We averaged the Insertion  score over all values of \(\alpha\) attempted and the removal scores over all values of \(\gamma\) tried and report layer-wise scores in Tables \ref{table:layer_down21}, \ref{table:layer_mid0}, \ref{table:layer_up00} and \ref{table:layer_up01}.
\begin{table}[t]
\centering
\caption{Removal FID and \(\frac{\text{UA}+\text{IRA}}{2}\) and insertion EPR at \texttt{down.2.1}, means over the 20 objects (EPR pooled over all edited images). Under \(\gamma_{\text{auto}}\) / \(\gamma_{\text{search}}\), EPR is insertion at \(\alpha = -\gamma\). In each column, the best method is in \textbf{bold} and the second best \textit{italicized}.}
\label{table:layer_down21}
\resizebox{\columnwidth}{!}{%
\begin{tabular}{|l|ccc|ccc|c|}
\hline
    & \multicolumn{3}{c|}{\(\gamma_{\text{auto}}\)} & \multicolumn{3}{c|}{\(\gamma_{\text{search}}\)} & \multicolumn{1}{c|}{\(\alpha_{\text{auto}}\)} \\
    Method & FID \(\downarrow\) & \(\frac{\text{UA}+\text{IRA}}{2}\) \(\uparrow\) & EPR \(\uparrow\) & FID \(\downarrow\) & \(\frac{\text{UA}+\text{IRA}}{2}\) \(\uparrow\) & EPR \(\uparrow\) & EPR \(\uparrow\) \\
\hline
    ESSI & \textit{9.897} & \textit{0.833} & \textit{1.570} & 10.431 & \textit{0.911} & \textbf{2.641} & 1.994 \\
    Adaptive & \textbf{9.660} & 0.830 & \textbf{1.760} & \textbf{8.902} & 0.845 & \textit{2.520} & \textit{2.481} \\
    OSP & 11.441 & 0.830 & 1.411 & \textit{9.274} & 0.804 & 1.882 & 1.860 \\
    SAeUron & 14.083 & \textbf{0.849} & 1.395 & 15.094 & \textbf{0.960} & 2.476 & \textbf{2.592} \\
    CASL (down) & 89.733 & 0.503 & 0.536 & 48.472 & 0.502 & 0.510 & 0.536 \\
\hline
\end{tabular}}
\end{table}

\begin{table}[t]
\centering
\caption{Removal FID and \(\frac{\text{UA}+\text{IRA}}{2}\) and insertion EPR at \texttt{mid.0}, means over the 20 objects (EPR pooled over all edited images). Under \(\gamma_{\text{auto}}\) / \(\gamma_{\text{search}}\), EPR is insertion at \(\alpha = -\gamma\). In each column, the best method is in \textbf{bold} and the second best \textit{italicized}.}
\label{table:layer_mid0}
\resizebox{\columnwidth}{!}{%
\begin{tabular}{|l|ccc|ccc|c|}
\hline
    & \multicolumn{3}{c|}{\(\gamma_{\text{auto}}\)} & \multicolumn{3}{c|}{\(\gamma_{\text{search}}\)} & \multicolumn{1}{c|}{\(\alpha_{\text{auto}}\)} \\
    Method & FID \(\downarrow\) & \(\frac{\text{UA}+\text{IRA}}{2}\) \(\uparrow\) & EPR \(\uparrow\) & FID \(\downarrow\) & \(\frac{\text{UA}+\text{IRA}}{2}\) \(\uparrow\) & EPR \(\uparrow\) & EPR \(\uparrow\) \\
\hline
    ESSI & 27.228 & 0.694 & \textit{0.751} & 13.596 & \textit{0.714} & 0.782 & \textbf{1.125} \\
    Adaptive & \textbf{12.875} & 0.710 & 0.716 & \textbf{9.786} & 0.699 & \textbf{0.847} & 0.919 \\
    OSP & \textit{14.022} & \textit{0.722} & 0.707 & \textit{10.613} & 0.702 & 0.821 & \textit{0.928} \\
    SAeUron & 42.497 & \textbf{0.741} & \textbf{0.762} & 35.365 & \textbf{0.811} & \textit{0.828} & 0.784 \\
    CASL (mid) & 171.545 & 0.486 & 0.637 & 79.391 & 0.496 & 0.625 & 0.637 \\
\hline
\end{tabular}}
\end{table}

\begin{table}[t]
\centering
\caption{Removal FID and \(\frac{\text{UA}+\text{IRA}}{2}\) and insertion EPR at \texttt{up.0.0}, means over the 20 objects (EPR pooled over all edited images). Under \(\gamma_{\text{auto}}\) / \(\gamma_{\text{search}}\), EPR is insertion at \(\alpha = -\gamma\). In each column, the best method is in \textbf{bold} and the second best \textit{italicized}.}
\label{table:layer_up00}
\resizebox{\columnwidth}{!}{%
\begin{tabular}{|l|ccc|ccc|c|}
\hline
    & \multicolumn{3}{c|}{\(\gamma_{\text{auto}}\)} & \multicolumn{3}{c|}{\(\gamma_{\text{search}}\)} & \multicolumn{1}{c|}{\(\alpha_{\text{auto}}\)} \\
    Method & FID \(\downarrow\) & \(\frac{\text{UA}+\text{IRA}}{2}\) \(\uparrow\) & EPR \(\uparrow\) & FID \(\downarrow\) & \(\frac{\text{UA}+\text{IRA}}{2}\) \(\uparrow\) & EPR \(\uparrow\) & EPR \(\uparrow\) \\
\hline
    ESSI & 17.678 & \textit{0.800} & \textbf{1.062} & 13.010 & \textit{0.846} & \textbf{1.264} & 1.165 \\
    Adaptive & \textbf{15.886} & \textbf{0.836} & 1.021 & \textbf{12.158} & 0.845 & \textit{1.185} & 1.154 \\
    OSP & \textit{17.430} & \textbf{0.836} & \textit{1.034} & \textit{12.307} & 0.823 & 1.081 & \textit{1.514} \\
    SAeUron & 19.575 & 0.768 & 0.751 & 20.776 & \textbf{0.942} & 1.115 & \textbf{1.538} \\
\hline
\end{tabular}}
\end{table}

\begin{table}[t]
\centering
\caption{Removal FID and \(\frac{\text{UA}+\text{IRA}}{2}\) and insertion EPR at \texttt{up.0.1}, means over the 20 objects (EPR pooled over all edited images). Under \(\gamma_{\text{auto}}\) / \(\gamma_{\text{search}}\), EPR is insertion at \(\alpha = -\gamma\). In each column, the best method is in \textbf{bold} and the second best \textit{italicized}.}
\label{table:layer_up01}
\resizebox{\columnwidth}{!}{%
\begin{tabular}{|l|ccc|ccc|c|}
\hline
    & \multicolumn{3}{c|}{\(\gamma_{\text{auto}}\)} & \multicolumn{3}{c|}{\(\gamma_{\text{search}}\)} & \multicolumn{1}{c|}{\(\alpha_{\text{auto}}\)} \\
    Method & FID \(\downarrow\) & \(\frac{\text{UA}+\text{IRA}}{2}\) \(\uparrow\) & EPR \(\uparrow\) & FID \(\downarrow\) & \(\frac{\text{UA}+\text{IRA}}{2}\) \(\uparrow\) & EPR \(\uparrow\) & EPR \(\uparrow\) \\
\hline
    ESSI & 55.938 & \textit{0.706} & \textit{0.864} & 31.472 & \textit{0.756} & \textbf{1.400} & \textbf{1.819} \\
    Adaptive & \textit{26.321} & \textbf{0.746} & 0.844 & \textbf{14.513} & 0.730 & 0.789 & 1.190 \\
    OSP & \textbf{23.615} & \textbf{0.746} & 0.816 & \textit{15.041} & 0.735 & 0.814 & 1.056 \\
    SAeUron & 81.633 & 0.614 & \textbf{0.897} & 59.617 & \textbf{0.798} & \textit{1.048} & \textit{1.590} \\
\hline
\end{tabular}}
\end{table}

\end{document}